# Persian Wordnet Construction using Supervised Learning

Zahra Mousavi
School of Electrical and
Computer Engineering,
College of Engineering,
University of Tehran,
Tehran, Iran
sz.mousavi@ut.ac.ir

Heshaam Faili
School of Electrical and
Computer Engineering,
College of Engineering,
University of Tehran,
Tehran, Iran
hfaili@ut.ac.ir

*Abstract*— **This paper presents an automated supervised method for Persian wordnet construction. Using a Persian corpus and a bi-lingual dictionary, the initial links between Persian words and Princeton WordNet synsets have been generated. These links will be discriminated later as correct or incorrect by employing seven features in a trained classification system. The whole method is just a classification system, which has been trained on a train set containing FarsNet as a set of correct instances. State of the art results on the automatically derived Persian wordnet is achieved. The resulted wordnet with a precision of 91.18% includes more than 16,000 words and 22,000 synsets.**

*Keywords- wordnet; ontology; supervised; Persian language*

## I. INTRODUCTION

Over the past years, acquiring semantic knowledge about lexical terms has been the concern of many projects in query expansion, text summarization [1], text categorization [2] and generating concept hierarchies [3]. For some languages such as English, broad coverage semantic taxonomy like Princeton WordNet (PWN) [4] has been constructed manually by spending great cost and time. Also, two great efforts in constructing wordnet for other languages were EuroWordNet [5] and BalkaNet [6]. The former deals with European languages such as English, Dutch, German, French, Spanish, Italian, Czech, and Estonian, and the latter deals with languages from Balkan area such as Romanian, Bulgarian, Turkish, Slovenian, Greek and Serbian.

A common feature among wordnets in different languages is synset. Synsets are sets of synonyms, which are connected together by means of semantic relations.

Two main strategies for automatically constructing wordnet can be considered: 1) Merge and 2) Expansion [5]. In the merge approach, an independent wordnet for target language is created, and for each synset in the generated wordnet equivalent synsets in PWN or another available wordnet is identified. This method is more complex than expansion approach and requires more time to construct a wordnet. The available lexical resources and wordnet building tools and also, the polysemy of the words in the synsets, directly affect the average time is consumed for building each lexical entry of wordnets. In the expansion approach, one available wordnet, usually PWN, is considered as source wordnet, and the words associated to its synsets are translated to the target language to generate the initial synsets of the wordnet. This process is based on an assumption, which implies that the concepts and their relations are language-independent, while it may be disaffirmed in some cases. Therefore, the coverage of language-specific concepts and properties isn't warranted by the produced wordnet, which is a drawback of the expansion approaches. In these approaches, the structure of the source wordnet is used for target language and other meta-data over source wordnet such as Domain models can be used for target wordnet, too. Consequently, it excludes time-consuming and expensive manual process for providing such information. The other advantage of this approach is automatic aligning wordnets to each other, which can be exploited in NLP multilingual tasks extensively. In general, the expansion approach is an efficient method for WordNet construction, but the generated wordnet is heavily biased or limited to the source wordnet.

In EuroWordNet and BalkaNet projects a top-down methodology has been used. In the first step of this methodology, a core wordnet has been developed manually which contains all high-level concepts of the language. At the next step, core wordnet has been expanded using automated techniques with high confident results. Using this approach, a number of automated methods were proposed for constructing a wordnet for Asian languages such as Japanese, Arabic, Thai, and Persian, which uses PWN and other existing lexical resources.

In recent years, some efforts have been made in order to create a wordnet for Persian language. In fact, different methods to construct Persian wordnet manually, semi-automatically and automatically have been proposed. In [7] a semi-automatic method is proposed in which for each Persian word, a number of PWN synsets is suggested by the system in order to be judged later by a human annotator to select a relevant synset. By using some other automated methods with human supervision, their work in construction of Persian wordnet has been expanded later, and an initial Persian wordnet named FarsNet has been developed [8]. In [9] an automatic method for Persian WordNet construction based on PWN is introduced. The proposed method uses a bi-lingual dictionary and Persian and English corpora to link Persian words to PWN synsets. A score function has been defined to rank the mappings between Persian words and PWN synsets. In the next work [10], a word sense disambiguation (WSD) method is employed in an iterative approach based on Expectation-Maximization (EM) algorithm to estimate a probability for each candidate synset linked to Persian words. Another iterative approach is presented in [11] in which the estimation of probabilities is performed based on Markov chain Monte Carlo algorithm. An extension of [10] is described in [12], which succeeded to improve the results by employing a graph-based WSD method. After execution of the EM algorithm, all links with a probability under a pre-determined threshold were removed from the wordnet. Considering 0.1 as the value of threshold acquired a wordnet composed of 11,899 unique words and 16,472 WordNet synsets with a precision of 90%. In this paper, we use this wordnet, the state-of-the-art automatically constructed Persian wordnet, as the baseline for evaluating our wordnet.

In this paper, an expansion-based approach is proposed for constructing a Persian wordnet. Most of previously proposed methods for automatically construction of Persian wordnet follow unsupervised approaches. We intend to present a supervised wordnet construction due to their higher accuracy in comparison with unsupervised methods. However, supervised methods usually suffer from the lack of sufficient reliable labeled data. In this research, a train dataset is produced by utilizing FarsNet, the pre-existing Persian wordnet. In fact, the main idea of this work is exploiting the available links between FarsNet and PWN synsets to link other Persian words to PWN synsets. Similar to the work of [13], the construction method is defined as a classifier. By defining seven features for each link, the classifier is able to classify the links into two categories: correct and incorrect. Available Persian resources are employed to extract distributional and semantic features. Also, the feature set is enriched by utilizing efficient methods for measuring lexical semantic similarity such as Word2Vec model [14]. Evaluation of the results indicates an improvement comparing to the previously built Persian wordnets.

The rest of the paper is organized as follows. Section 2 presents an overview on some automated methods proposed for constructing wordnets. Section 3 presents our method for automatically extending the Persian wordnet. Experimental results and evaluation of the proposed method are explained in Section 4. Finally, conclusion and future works are presented in Section 5.

## II. RELATED WORKS

Many researchers have proposed different approaches for automatically constructing wordnets. In [13] an automatic method for construction of a Korean wordnet using PWN has been presented. In this work, links between Korean words and PWN synsets have been made using a bi-lingual dictionary. These links are classified as correct or incorrect by using a classifier with six features, which is trained on a set containing 3260 manually classified instances. The performance of each feature has been examined by means of precision and coverage as the proportion of linked senses of Korean words to all the senses of Korean in a test set. The best feature had 75.21% precision and 59.5% coverage. In addition, the experiments have shown that the precision for each features, is always better than random choice baseline. The combination of features using decision tree showed 93.59% precision and 77.12% coverage for Korean language.

In [15] the basic English-Russian wordnet based on the English-Russian lexical resources and morphological analyzer tools was built. Also, in [16] a pattern-based algorithm for extracting lexical-semantic relations in Polish is presented.

In [17], an effort has been done for extending Arabic wordnet using lexical and morphological rules and applying Bayesian inference in semi-automatic manner. In this research in order to associate Arabic words with PWN synsets, a Bayesian network with four layers has been proposed. In the first layer, Arabic words have been located and their corresponding English translations are placed in the second layer. All the synsets of English words existing in layer 2, have been set in layer 3. Layer 4 is additional layer of PWN synsets, which has been associated with the synsets of layer 3 by way of semantic relation. For the Arabic words with only one English translation, which this translation is monosemous, too and moreover for the Arabic words with English translations belonging to a common synset, association between the words and the common PWN synset have been made directly. In other cases a learning algorithm has been applied for measuring the reliability of each <Arabic word, PWN synset> association. A set of candidates is built with pairs <X, Y> where X belongs to Arabic words and Y belongs to PWN synsets in layer 3 of Bayesian network and has a non-zero probability, also there is a path from X to Y. The tuple is scored with the posterior probability of Y given the evidence provided by the Bayesian network. Only the tuples scored over a predefined threshold were selected for inclusion in the final set of candidates. The best result obtained from the mentioned method in this research showed precision of 71%.

By examining candidate synsets of a given word in target language and their relations, some criteria can be defined, which represent some features of correct links. In [18], such idea for constructing Thai wordnet has been proposed. They defined 13 criteria, which have been categorized into three groups: Monosemic criteria which focus on English words with only one meaning, Polysemic criteria which focus on English words with multiple meanings and Structural criteria which focus on the structural relations between candidate synsets. In order to verify the constructed links using these 13 criteria, stratified sampling technique has been applied. The results of verification showed 92% correctness for the best criterion and 49.25%. was reported as the lowest correctness.

In [7], a Persian core wordnet was constructed for a set of common base concepts. In order to extend the core wordnet, for each synset in PWN, all Persian translations of English words were extracted using a bi-lingual dictionary and the appropriate translations were identified using two heuristics and a WSD method. The manual evaluation of the resulted links between Persian words and PWN synsets showed precision of about 72% in the resulting Persian lexicon. This work was extended in [8] and published as the first Persian wordnet, called FarsNet. Three methods for extracting conceptual relations for nouns were presented. In the first method, a set of 24 patterns to extract taxonomic relations has been defined. While in the second approach, Wikipedia page structures such as tables, bullets, and hyperlinks have been used to extract some relations between word pairs. Finally in the third method, morphological rules have been applied on a corpus to extract antonymy relations between adjectives. Their system employs linguistic and statistical methods to cluster adjectives. Adjectives that defined different degree of the same attribute are put in one cluster.

In [9] an automatic method for Persian wordnet construction based on PWN is introduced. It uses a score function for ranking the mappings between Persian words and PWN synsets, and the final wordnet is built by selecting the highest scores. In the next work [10], they proposed an unsupervised method using EM algorithm to construct a Persian wordnet. In order to determine candidate synsets for each Persian word, a bi-lingual dictionary and PWN were utilized. Next, a probability was calculated for each candidate synset applying a WSD method in Expectation step. These probabilities were being updated in each iteration of EM algorithm until convergence to a steady state. Finally, a wordnet including 7,109 unique words and 9,427 PWN synsets, was adopted by extracting 10% of high probable word-synset pairs. The evaluations showed a precision of 86.7% according to a manual test set consists of about 1,500 randomly selected word-synset pairs. An extension of this work is described in [12], which succeeded to improve the results by changing the WSD method. Also, this method is applicable to low-resource languages due to the employed resources. The resulted wordnet consists of 11,899 Persian words and 16,472 PWN synsets with about 30,000 word-synset pairs, gained a score of 90% with respect to precision.

A similar iterative approach using Markov chain Monte Carlo algorithm was presented in [11] to construct a Persian wordnet. This method approximates the probabilities of each candidate synset assigned to Persian words based on a Bayesian Inference. Selecting 10,000 word-synset pairs with highest probabilities, resulted to a wordnet with the precision of 90.46%.

III. PERSIAN WORDNET CONSTRUCTION

The proposed method uses Princeton WordNet, a bi-lingual dictionary, a pre-existing Persian wordnet, FarsNet, and a Persian corpus as its available resources. Each concept in English is represented by one synset in PWN. Based on the assumption of the Expansion method, it is considered that for the most concepts in English, there exists an equivalent concept in Persian and the language-specific concepts are ignored. Thus, by identifying the proper translations of an English word appearing in each synset, a Persian synset representing the same concept as the English one can be constructed.

Bijankhan Persian corpus [19] is employed as the resource for extracting Persian words of the wordnet. It leads to coverage of more frequently used Persian words in the resulting wordnet. Bijankhan corpus is available in two versions, which the second release is used in our experiments. It is a collection of daily news and common texts. All documents in this collection are grouped into about 4300 different subject categories. This corpus contains about ten millions manually tagged words with a tag set including 550 Persian part of speech (POS) tags [20].

The first step for wordnet construction is translating the Persian words by a bi-lingual dictionary to English counterparts. But before translating the words, it's necessary to employ a lemmatizer tool to adapt the different forms of the words. Otherwise, some words existing in the corpus may not be detected in the dictionary due to appearing in the inflection forms. In this regard, STeP-1 [21] tool is exploited. It contains some Persian text processing tools such as tokenizer, spell checker, morphological analyzer and POS tagger.

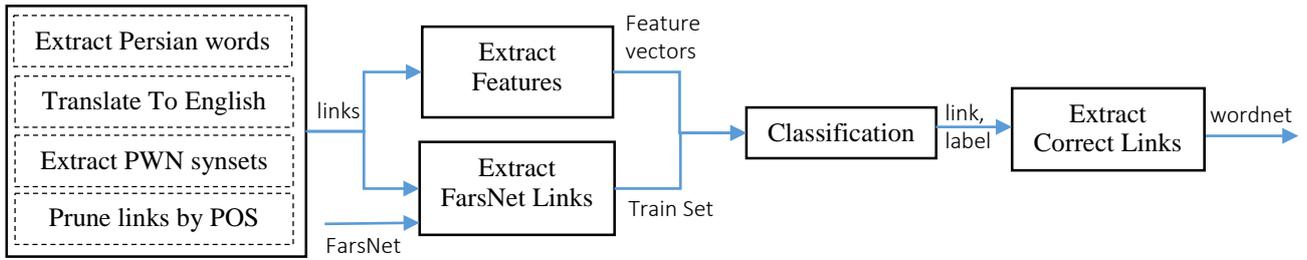

*Figure 1: the Overview of proposed methods for construction of Persian wordnet*

Next, each lemmatized Persian word is translated to English equivalents by Aryanpour [1] Persian to English dictionary. Then Princeton WordNet 3.0 is used to identify English candidate synsets for each Persian word. Determining all the PWN synsets including English translations of a Persian word, the initial links between that Persian words and PWN synsets are generated. It is possible that more than one Persian word linked to the same PWN synset. Because of English word polysemy, some of these Persian words don't imply the same meaning as the meaning of their linked synset. In fact, there are several invalid links between Persian words and PWN synsets, which should be removed. Some of these links can be deleted by exploiting extra knowledge about Persian words. As mentioned, Bijankhan corpus is enriched by POS tagging. This corpus gives proper evidence about POS tags of each Persian word. By using this corpus, the probability of observing each Persian word with each POS tag of noun, verb, adjective and adverb is calculated. This information is used to eliminate incompatible links between PWN synsets and Persian words. The incompatible link is the one that is made between a PWN synset and a Persian word with inconsistent POS tags. Consequently, 47,291 links out of 247,947 links are pruned and totally 200,656 candidate links are remained. However, there are still many false links, which must be removed. For this purpose, seven features for each of these links have been introduced. Using these features, a classifier to discriminate these links as correct or incorrect links has been trained. To define some of these features, some measures of corpus-based semantic similarity and relatedness have been used.

Over the past years, many articles addressed the notion of lexical semantic similarity [22]. The studies in this field attempted to determine how two words are semantically close and what semantic relation they share, if similar. Another field that is even more general than semantic similarity is semantic relatedness [22]. In this area some efforts have targeted designing similarity measures that exploit more or less structured source of knowledge such as WordNet, dictionaries, Wikipedia articles and corpora. Most of these measures are defined based on distributional hypothesis, which is based on the idea that words found in similar context have more chance to be similar. Each word in the corpus is characterized with a context vector. Each element of this vector is considered as a feature and its value is calculated by lexical association measures. Semantic similarity between two words is then calculated by computing similarity measures on context vectors of each given word pair.

In our experiments context vector of Persian words was constructed using Bijankhan corpus. In this study co-occurrence frequency for extracting context vector of each word from the corpus has been used. Contexts were restricted to the words within the sentence containing the target word and one hundred words, which have the highest co-occurrence frequency with each word in the context, are considered as the context vector (CV) of that word.

Recently, neural embedding techniques such as Word2Vec [14] have attracted lots of attention of researches. Word2Vec is an unsupervised method for learning distributional real-valued representations of words by using their contexts to capture the relation between the words. Due to its effectiveness, It has been widely used in many Natural Language Processing (NLP) tasks since its publication. Indeed, it transfers the words to a low-dimensional vector space, which is able to represent the words with similar contexts properly in a close proximity of the space. Hence, it gives a good metric for semantically comparing the words by using vector-based similarity measures.

In our experiments by exploiting Word2Vec model, 300-dimensional vectors for Persian words have been trained using Bijankhan corpus. Using these vectors, semantic similarity between each pairs of Persian words can be computed. Here Cosine similarity measure was used to calculate similarity between two words.

Similar to the procedures carried out for Persian words, in the case of English words, about 500 megabytes of English Wikipedia documents were considered and a context vector for each English word was constructed.

---

[1] See http://www.aryanpour.com

As mentioned the whole method is just a classifier system, which has been trained on a generated training data set. By employing seven features in this classifier the links between Persian words and PWN synsets are classified into two distinct categories: correct and incorrect. The final Persian WordNet is a set of all links, which have been classified as correct links. Figure 1 illustrates an overview of the proposed methods for wordnet construction.

We used the links between Persian words and PWN synsets, which have been presented in FarsNet as correct instances of training data. Also, a set of randomly selected links were added to training data as incorrect instances. By exploiting distributional and semantic information extracted from available Persian resources, seven features for the classification task have been defined which are described in the following subsections.

*A. Relatedness Measure*

In [9] a measure for calculating the relatedness measure between PWN synsets and Persian words has been defined. One of the drawbacks of the mentioned measure is the usage of path WordNet similarity. This similarity measure has the restriction, which is only applicable to nouns and verbs. Here another approach is used to define a new relatedness measure for each link. One of the basic ideas for calculating semantic similarity between two words is based on this fact that two words are similar if their context vectors be similar [22]. So, in the case of English words appearing in the same synset, it's expected that they appear in the same context and thus have similar context vector. Based on the above notion, a relatedness measure between an English word and a PWN synset can be defined using formula 1.

$$Relatedness(e, s) = \frac{\sum_{e' \in s} \frac{|CV(e) \cap CV(e')|}{|CV(e) \cup CV(e')|}}{|\{e' | e' \in s\}|} \quad (1)$$

Where the |.| operator gives the size of given collection.

According to this formula, an English word $e$ has the highest relatedness with respect to a PWN synset $s$ if it is a related word of all words appeared in synset $s$. As previously mentioned, context vector of each Persian word was extracted from a corpus. Using Aryanpour Persian to English dictionary, equivalent English translations of these words were extracted which called context vector translation (CVT). By considering the link between a PWN synset $s$ and a Persian word $f$, this inference can be made that if $f$ implies the same concept as $s$ then its context vector is more similar to the context vector of words in $s$. Because the words in $s$ are in English and $f$ is in Persian, CVT of Persian word was used to calculate this similarity. Thus the relatedness measure of the link between $f$ and $s$ is high if CVT members have high relatedness respect to $s$. However, this possibility must be taken into account that despite the high relatedness of a CVT element $e$ with $s$, there might be other senses of words within $s$ which $e$ has higher relatedness to them. Therefore, the relative relatedness of $e$ and $s$ to the summation of relatedness between $e$ and all synsets containing words of $s$ is considered rather than the relatedness of $e$ and $s$, itself. According to the following formula, the average of relative relatedness of CVT elements and $s$ is computed as relatedness measure (R) of $f$ and $s$.

$$R(f, s) = \frac{\sum_{e \in CVT} \frac{Relatedness(e, s)}{\sum_{s'} Relatedness(e, s')}}{|CVT|} \quad (2)$$

Where $s'$ is the member of all PWN synsets, which contains the English words appeared in $s$ and *Relatedness* is calculated using formula 1.

Since this feature isn't computable for the Persian words without context vector, the English equivalents of Persian word $f$ which links it to PWN synset $s$ can be considered as CVT too.

*B. Synset Strength*

The second feature is based on the idea that if two words are synonym then they usually appear in the same context [22]. As previously mentioned, the basic method for discovering synonym words is finding the words that have similar context vector. Persian words, which have been correctly linked to a PWN synset, are more probable to be synonym. Thus, their representative vectors must be similar. Consider $k$ Persian words $f_1, f_2, f_3, ..., f_k$ which linked to same PWN synset $s$. For Persian word $f$ and PWN synset $s$, Synset Strength (SS) feature is set to one in the case of $k=1$ and otherwise it is defined as follows:

$$SS(f, s) = \frac{\sum_{i=1, f_i \neq f}^{k} p(f_i, s) \times Similarity(f, f_i)}{k - 1} \quad (3)$$

Where $p(f_i, s)$ is the summation of the inverse of polysemy degree of English words which link Persian word $f_i$ to PWN synset $s$. The Similarity measure between two Persian words $f_i$ and $f_j$ is calculated by computing Cosine similarity measure on the vectors trained by Word2Vec model.

*C. Context Overlap*

A general definition or example sentence has been provided in PWN for each synset. One of the basic algorithms for word sense disambiguation (WSD) task is Lesk approach [23]. This algorithm uses dictionary definitions pertaining to the various senses of the ambiguous words in order to identify the most likely meanings of the words in a given context. This idea is used here to rate various Persian translations of each PWN synset. In order to disambiguate Persian translations of each PWN synset, the overlap between context vector of Persian word and Persian translation of the words in PWN synset gloss is considered. This feature is calculated using formula 4.

$$ContextOverlap(f, s) = \frac{|GT(s) \cap CV(f)|}{|GT(s) \cup CV(f)|} \quad (4)$$

Where $GT$ represents the set of Persian translations of gloss words in PWN synset $s$.

## D. Domain Similarity

Another similarity measure was defined here between two Persian words that exploits domain categories of documents in Hamshahri text corpus. Hamshahri is one of the online Persian newspapers in Iran, which has been published for more than 20 years and its archive has been presented to the public. In [24] this archive has been used and a standard text corpus with 318,000 documents containing about 110 million words has been constructed. The documents in this corpus have been categorized into nine main categories and 36 subcategories (like Economy, Economy. Bourse, …). For each Persian word $f$, a 9-dimensional vector was considered, one element for each category, as domain distribution of $f$. The value of $i$-th element is defined as the probability of occurring Persian word $f$ in the documents of $i$-th category.

Domain similarity between two Persian words is calculated by using the Jensen-Shannon divergence, which is a popular method of measuring the similarity between two probability distributions. The square root of the Jensen–Shannon divergence is a metric often referred to as Jensen-Shannon distance [25, 26]. The Jensen-Shannon divergence between two distributions P and Q is calculated using formula 5.

$$JS(P,Q) = \frac{1}{2}(D(P \| M) + D(Q \| M)) \quad (5)$$

The function D is the Kullback-Leibler divergence, and M is the average of P and Q. Formula 6 is used to compute the similarity between two distributions P and Q.

$$Similarity(P,Q) = 1 - \sqrt{JS(P,Q)} \quad (6)$$

Domain Similarity measure is based on the idea that it is expected that synonym words appear in the same domain or the distribution of synonym words in different domains is similar. So, according to this feature a link between a Persian word $f$ and a PWN synset $s$ is correct if $f$ appears in the same domain as the domains that other Persian words linked to synset $s$, appear in. If just one Persian word $f$ is linked to PWN synset $s$, the value of this feature for the corresponding link will be set to one. Now, consider Persian words $f_1$, $f_2, f_3, …, f_k$, which are linked to same PWN synset $s$. For Persian word $f$ and PWN synset $s$, Domain Similarity (DS) is defined as follows:

$$DS(f,s) = \frac{\sum_{i=1, f_i \neq f}^{k} p(f_i, s) \times Similarity(D_f, D_{f_i})}{k-1} \quad (7)$$

Where $p(f_i, s)$ is the summation of the inverse of polysemy degree of English words which link Persian word $f_i$ to PWN synset $s$ and $D_f$ is the domain distribution of Persian word $f$.

## E. Monosemous English

This feature is similar to the first heuristic defined in [7]. Suppose that word $e$ is an English translation of Persian word $f$. If there has been only one synset $s$ in PWN that contains $e$ as a member, then the value of this feature for the link between $f$ and $s$ is set to one and in the other case, zero. Since $e$ is an English translation of $f$, it shares some concepts with $f$. So, there are some senses of $e$ in PWN that have equivalent concept with Persian word $f$. In the case that English word $e$ appears in one synset, we suppose that this synset implies the common concept with Persian word $f$ and set the value of this feature to one. It should be considered that it is possible that Persian word $f$ may have more than one sense, which will be proposed with its other English translations.

## F. Synset Commonality

This feature has been defined similar to the second heuristic defined in [7]. This feature shows the number of different English words that link a Persian word $f$ to a PWN synset $s$. Whatever more English translations suggest a PWN synset $s$ for a given Persian word $f$, it is more probable that common meaning between $f$ and its English translations be synset $s$. Thus if Persian word $f$ has several English translations and there is a PWN synset that has $m$ of those English translations as member then the value of this feature is set to $m$.

## G. Importance

In a Persian/English dictionary, different meanings of each Persian word can be represented by different English words. On the other hand, for each English word, one or more senses have been presented in PWN. With this assumption that each English translation of a given Persian word represents one of its meaning, for each English translation one of its senses has the same meaning with Persian word. The Importance feature was defined to exploit this assumption. The value of this feature was calculated using values of other features. Consider Persian word $f$ and one of its English translations $e$. Suppose $s_1, s_2, …, s_k$ are synsets in PWN, which contain $e$ as their member. The Importance feature for a link between $f$ and $s_i$ is calculated as follows: four features, Relatedness measure, Synset Strength, Context Overlap, and Domain Similarity, are initially taken into consideration. For each of which, if $s_i$ has the maximum value compared to the other synsets of English word $e$ then Importance value of link between $f$ and $s_i$ is increased by one. In fact, the link between Persian word $f$ and PWN synset $s_i$ will have the highest Importance only if the value of the aforesaid features is the maximum, comparing to the other synsets of English word $e$.

## IV. EXPERIMENTS AND RESULTS

The goal of the experiments is to assess the effectiveness of the proposed features in discriminating between correct and incorrect links by evaluating the accuracy of classification system. As mentioned, the approach is to train a classifier that makes use of these features. In order to train such classifier, we need a collection of classified links as training set. In this regard, we considered the usage of pre-existing Persian wordnet, FarsNet, which is the first published Persian WordNet. The process of building train data relies on the second release of FarsNet. This version organizes more than 36,000 Persian words and more than 20,000 synsets in different hierarchical structures. It also contains inter-lingual relations connecting Persian synsets to English synsets of Princeton WordNet 3.0. Taking advantage of these links, we are able to obtain correct instances of train data. Table 1 shows some statistics about FarsNet 2.0. For each available link between Persian words and PWN synsets such as *(f, s)* in FarsNet, an instance *(f, s, correct)* was considered as correct instance of training set.

| Category | Words | Synsets | Links to PWN |
|---|---|---|---|
| Noun | 22,180 | 11,954 | 10,108 |
| Adjective | 6,560 | 4,261 | 4,516 |
| Adverb | 2,014 | 923 | 929 |
| Verb | 5,691 | 3,294 | 2,678 |
| Total | 36,445 | 20,432 | 18,231 |

Table 1: Statistics of FarsNet 2.0

By considering the whole available links in FarsNet, 10,952 links are added to training set as correct class. In order to generate incorrect instances of training set, 5,000 links between Persian words and PWN synsets excluding FarsNet links, were selected randomly and added to training set as *(f, s, incorrect)*. In general a train set consists of 10,952 correct and about 5,000 incorrect instances, was obtained. Due to overlap of some links with the gold dataset, that is used in the evaluation process of experiments, several links were eliminated. The statistics of the final training set is reported in Table 2.

| POS | Correct | Incorrect | Total |
|---|---|---|---|
| Noun | 7,974 | 3,288 | 11,262 |
| Adjective | 2,357 | 1,261 | 3,618 |
| Adverb | 217 | 82 | 299 |
| Verb | 316 | 363 | 679 |
| Total | 10,864 | 4,994 | 15,858 |

Table 2: Statistics of train set

For each of links in training set, defined features were calculated. In our experiments, Weka open source data mining software [27] was used. In order to evaluate the classifier accuracy, two methods were considered. The first method uses ten-fold cross validation testing method provided by Weka. Table 3 shows the precision and recall measures obtained from different classifiers. Because the final Persian wordnet is generated by collecting the links classified as correct, the precision of correct class instances is more important than the other measures. The last two columns of Table 3 show the precision and recall measures of correct class, with respect to different classifiers: Random Forest, KNN, Multilayer Perceptron, and Naïve Bayes.

| Classifier | Precision | Recall | Correct Precision | Correct Recall |
|---|---|---|---|---|
| NaïveBayes | 70.4 | 58.3 | 83.6 | 48.6 |
| KNN (k=10) | 67.7 | 69.9 | 74.2 | 86.1 |
| RandomForest | 67.9 | 70.2 | 74.2 | 86.5 |
| MultilayerPerceptron | 68.0 | 70.6 | 73.3 | 89.6 |

Table 3: Precision and Recall of applied classifiers

As shown in Table 3, the best accuracy with respect to the precision of correct class was achieved by Naïve Bayes classifier. Therefore, Naïve Bayes classifier is employed to construct the final wordnet. The links classified as *correct class* excluding the existing links in FarsNet, were collected to make the final Persian wordnet with precision score of 83.6%.[2]

In order to assess the effect of each feature on the resulted wordnet, Naïve Bayes classifier is learned by different configuration of features. For this purpose, the worth of each feature is evaluated by measuring the information gain of each feature using Weka attribute selection. Next, features are incrementally added to the feature set in order of their information gain and the output of each step is given to a classifier. Table 4 shows the results of classifiers in terms of precision, recall and F-measure scores with respect to the correct class. Features are presented in this table according to the information gain rank.

| Features | Precision | Recall | F-measure |
|---|---|---|---|
| Importance | 68.5 | 100 | 81.3 |
| + Synset Commonality | 80.7 | 58.4 | 67.8 |
| + Relatedness Measure | 82.8 | 52.4 | 64.2 |
| + Domain Similarity | 82.6 | 50.4 | 62.6 |
| + Synset Strength | 83.4 | 48.2 | 61.1 |
| + Monosemous English | 83.7 | 48.4 | 61.3 |
| + Context Similarity | 83.6 | 48.6 | 61.5 |

Table 4: The results gained by classifiers trained on incrementally increasing feature set

As shown in Table 4, the precision measure is usually increasing as features are added. In some cases such as adding *Context Similarity* feature, precision falls down, while recall increases. Employing all the features leads to a precision of 83.6% and a recall of 48.6% according to ten-fold cross validation testing method.

Similar to other works in the PWN synset mapping, a manually judged test set is employed for evaluating the final links between Persian words and PWN synsets. In this regard, the method introduced in [12] is used as baseline. In this work as in our method, the initial links were generated by linking Persian words in Bijankhan corpus to PWN synsets. Next, an unsupervised EM-based algorithm using a cross-lingual WSD method has been applied to estimate the probabilities for each link. The final wordnet contained total links excluding low rated ones, which don't meet a pre-determined threshold. The highest precision in the experiments was gained by 0.1 as the threshold, which indicates a precision score of 90% and a recall of 35%. We address this wordnet as "EM-based wordnet" in contrast to our final wordnet as "Supervised wordnet". In the experiments of EM-based wordnet, a set of manually judged links has been obtained to evaluate the results. A subset of manual judges consists of about 1000 links corresponds to our generated links. Moreover, they aren't presented in the built training set. Therefore, we used this collection as test set in the evaluation process of the generated wordnet. Table 5 demonstrates some statistics about test dataset with respect to POS category and label.

---
[2] The resulted Persian WordNet is freely downloadable from http://ece.ut.ac.ir/en/node/940

| POS | Correct | Incorrect | Total |
| --- | --- | --- | --- |
| Noun | 440 | 109 | 549 |
| Adjective | 181 | 57 | 237 |
| Adverb | 27 | 4 | 31 |
| Verb | 103 | 84 | 187 |
| Total | 751 | 254 | 1,005 |

Table 5: Statistics of test set

Similar to [12] the precision is considered as the number of correct links are common in the wordnet and test data, divided by the total number of wordnet links which belong to the test data. Also, the recall of the wordnet is considered as the number of correct links are common in the wordnet and test data, divided by the total number of correct links in the test set.

The manual evaluation on the selected links shows a precision score of 91.18% and a recall score of 45.41%, which surpasses the EM-based wordnet, the state of the art automatically constructed Persian wordnet. Table 6 demonstrates the precision and recall of the supervised wordnet for different POS categories. The best precision was acquired for nouns with a score of 93.69% and the best recall dedicated to adverbs with a score of 51.85%.

| POS | Precision | Recall | F-measure |
| --- | --- | --- | --- |
| Noun | 93.69 | 47.27 | 62.84 |
| Adjective | 90.43 | 46.96 | 61.82 |
| Adverb | 93.33 | 51.85 | 66.67 |
| Verb | 79.07 | 33.01 | 46.58 |
| Total | 91.18 | 45.41 | 60.62 |

Table 6: Precision and Recall of resulted wordnet with respect to POS category

In addition to precision measure, the other noticeable factor for deliberating the quality of wordnets is their size. It denotes the number of unique words, synsets and word-sense pairs, covered by the wordnet. Table 7 represents this information about the induced wordnet.

The resulted wordnet covers about 16,000 words and 22,000 synsets and makes about two times more connections from Persian words to PWN synsets, in comparison with FarsNet. According to the first column of Table 7, nouns have the largest proportion of the resulted wordnet and the lowest coverage returns to verbs.

| POS | Words | Synsets | Word-sense Pairs |
| --- | --- | --- | --- |
| Noun | 10,486 | 13,947 | 23,425 |
| Adjective | 4,775 | 5,433 | 11,037 |
| Adverb | 460 | 508 | 778 |
| Verb | 408 | 2,883 | 3,107 |
| Total | 16,129 | 22,771 | 38,347 |

Table 7: Number of words, synsets and word-sense pairs in resulted Persian wordnet

In the following, the scalability of two wordnets from the perspective of the number of unique words, synsets and word-sense pairs, is studied. Table 8 reports these statistics for the induced wordnet and baseline method. Also, the number of unique words with more than one sense inside the wordnet, divided by the total number of unique words is represented in the last column of this table as polysemy rate. The higher polysemy rate in wordnets can be considered as a point of strength for them, due to leading more efficiency in NLP and IR tasks.

According to Table 8, supervised wordnet outperforms EM-based wordnet in respect of size, too. But the proportion of polysemic words, words with more than one sense, in EM-based wordnet is more than supervised wordnet.

|  | Unique Words | Synsets | Word-sense pairs | Polysemy rate |
| --- | --- | --- | --- | --- |
| EM-based wordnet | 11,899 | 16,472 | 29,944 | 0.73 |
| Supervised wordnet | 16,129 | 22,771 | 38,347 | 0.51 |

Table 8: Size of supervised wordnet in comparison with EM-based wordnet

The other measure considered in the evaluation of EM-based wordnet, regards to the coverage of Persian corpus words, PWN synsets and core concepts. Core concepts imply more frequently used synsets in a language, which covering them in a wordnet boosts its efficiency. A set of approximately the 5,000 most frequently used PWN word senses is created in [28], which is exploited here[3]. Table 9 compares supervised wordnet and EM-based wordnet from the coverage point of view. It's obvious that supervised wordnet has a wider coverage over Bijankhan corpus and PWN synsets, but EM-based wordnet has covered a higher percentage of core concepts.

|  | Bijankhan (unique words) | PWN synsets | Core synsets |
| --- | --- | --- | --- |
| EM-based wordnet | 11,543 | 14% | 53% |
| Supervised wordnet | 14,797 | 19.35% | 38.76% |

Table 9: Coverage of supervised wordnet in comparison with EM-based wordnet

---

[3] See http://wordnetcode.princeton.edu/standoff-files/core-wordnet.txt

In general, the experiments showed that supervised wordnet performed better than EM-based wordnet in many aspects. From the best of our knowledge, the retrieved precision is the highest accuracy comparing to whole other automatically built Persian wordnets. Also, it is the largest fully automatically constructed Persian wordnet, which covers more than 16,000 words, 22,000 PWN synsets and 38,000 word-sense pairs.

## V. Conclusion and Future Works

Automatic construction of Persian wordnet using available resources such as Persian and English monolingual corpora, bi-lingual dictionary, and Persian part of speech tagged corpus is the main concern of this paper. Also, FarsNet, the pre-existing Persian wordnet was exploited to produce a training set. For each link between Persian words and PWN synsets, seven features were defined and a classifier was trained to discriminate between correct and incorrect links. The features were defined by using measure of corpus-based semantic similarity and relatedness. Our experiments on Persian language showed the precision of 91.18% for the links that are classified as correct, which outperforms the previously proposed automated methods.

The experiments revealed that there are problems for calculating some features values. In PWN for some synsets a short gloss has been provided which causes the calculated Context Overlap feature for linked Persian words to be lower than other synsets that linked to those Persian words. In order to overcome this problem, synsets that have semantic relation with these synsets such as hypernyms can be considered.

Another observation that we made is that corresponding PWN synsets of some senses of English words contain only one English word. For example "bank" appears in 10 different noun synsets such that 6 of them contain only "bank". In these cases, the values of Synset Strength and Domain Similarity features become equal for all of links that derived from such English words. As we examined PWN, we observed that PWN contains 7,935 English words, which appear alone in more than one synset. This number of words is 5 percent of all English words in PWN and it is expected in these cases that other features discriminate between correct and incorrect links.

The experiments showed that verbs have the lowest proportion of the induced wordnet. Persian verbs are categorized into simple and compound verbs. Compound verbs are composed of a verbal and one or several non-verbal parts. This category of verbs includes a larger amount of Persian verbs. Since in the proposed method, Bijankhan corpus was used to extract Persian words and each token was specified as a single word, the extracted verbs usually correspond to simple verbs and our wordnet lacks a satisfactory coverage on compound verbs. We need a method for extracting the compound verbs from corpus, which can be considered as a future work. Also, the features can be enriched by POS wise features to have more accurate results.

The whole method is language-independent and can be experimented on each language whose needed resources are available.